\documentclass[10pt,twocolumn,letterpaper]{article}

\usepackage{iccv}              % To produce the CAMERA-READY version
\usepackage{stfloats}
\usepackage{expl3}
\usepackage{enumitem} 
\usepackage{multirow}
\usepackage[utf8]{inputenc}
\usepackage[accsupp]{axessibility}  % Improves PDF readability for those with disabilities.

\usepackage[normalem]{ulem}
\usepackage{tabularray}

\definecolor{myred}{RGB}{255, 99, 71}   
\definecolor{mygreen}{RGB}{60, 179, 113} 
\definecolor{myblue}{RGB}{30, 144, 255}

\definecolor{iccvblue}{rgb}{0.21,0.49,0.74}
\usepackage[pagebackref,breaklinks,colorlinks,allcolors=iccvblue]{hyperref}

\def\paperID{12318} % *** Enter the Paper ID here
\def\confName{ICCV}
\def\confYear{2025}

\title{GeoDistill: Geometry-Guided Self-Distillation for Weakly Supervised Cross-View Localization}
\author{Shaowen Tong\textsuperscript{1} \qquad Zimin Xia\textsuperscript{2} \qquad Alexandre Alahi\textsuperscript{2} \qquad Xuming He\textsuperscript{1} \qquad Yujiao Shi\textsuperscript{1}\thanks{Corresponding author.} \\
\textsuperscript{1}ShanghaiTech University, China \quad \textsuperscript{2} {\'E}cole Polytechnique F{\'e}d{\'e}rale de Lausanne (EPFL), Switzerland\\
{\tt\small \{tongshw2024,hexm,shiyj2\}@shanghaitech.edu.cn, \{zimin.xia,alexandre.alahi\}@epfl.ch}
}
\begin{document}
\maketitle

\begin{abstract}
Cross-view localization, the task of estimating a camera's 3-degrees-of-freedom (3-DoF) pose by aligning ground-level images with satellite images, is crucial for large-scale outdoor applications like autonomous navigation and augmented reality. Existing methods often rely on fully supervised learning, which requires costly ground-truth pose annotations. In this work, we propose GeoDistill, a \textbf{Geo}metry guided weakly supervised self \textbf{Distil}lation framework that uses teacher-student learning with Field-of-View (FoV)-based masking to enhance local feature learning for robust cross-view localization.
In GeoDistill, the teacher model localizes a panoramic image, while the student model predicts locations from a limited FoV counterpart created by FoV-based masking. By aligning the student's predictions with those of the teacher, the student focuses on key features like lane lines and ignores textureless regions, such as roads. This results in more accurate predictions and reduced uncertainty, regardless of whether the query images are panoramas or limited FoV images.
Our experiments show that GeoDistill significantly improves localization performance across different frameworks.  
Additionally, we introduce a novel orientation estimation network that predicts relative orientation without requiring precise planar position ground truth. GeoDistill provides a scalable and efficient solution for real-world cross-view localization challenges. Code and model can be found at \href{https://github.com/tongshw/GeoDistill}{https://github.com/tongshw/GeoDistill}.

% Despite advances in fully supervised cross-view localization, generalization to unseen cities remains limited by costly precise GPS annotations and significant performance drops. Inspired by robust human navigation using geometric cues, we introduce GeoDistill, a novel weakly supervised paradigm. GeoDistill pioneers geometry-guided self-distillation, leveraging noisy GPS and the inherent geometric consistency between ground and satellite views as a powerful inductive bias for learning robust 3-DoF pose estimation. Our teacher-student architecture compels a limited Field-of-View student to learn enhanced geometric matching from a full-panorama teacher, boosting cross-area generalization.  An Uncertainty preservation Strategy further ensures stable distillation. Validated by retraining state-of-the-art methods, GeoDistill achieves significant cross-area performance gains (18.5\% and 13.6\%) without architectural changes, demonstrating effective geometry-guided self-distillation for generalizable cross-view localization under weak supervision.
\end{abstract}    
\section{Introduction}
\label{sec:intro}
Visual localization estimates a camera's pose by matching its image to a known environment. Cross-view localization, specifically, determines the 3-degrees-of-freedom (3-DoF) camera pose, i.e., planar position and yaw orientation, by aligning a ground-level query image with a satellite image that covers its surroundings. 
Leveraging the widespread availability of satellite images, this approach has attracted significant attention for large-scale outdoor applications, including autonomous navigation and augmented reality.
% Visual localization, the task of determining a camera's precise pose within a known environment, is a fundamental capability underpinning a wide range of applications, including autonomous navigation, augmented reality, and robotics.  
% Among its variants, cross-view localization has garnered significant attention, particularly for large-scale outdoor scenarios.
% This approach leverages geo-referenced satellite or aerial imagery as a map representation, enabling the estimation of a ground-level camera's 3-degrees-of-freedom (3-DoF) pose – planar position and yaw orientation – relative to the overhead view. The widespread availability of publicly accessible satellite imagery with global coverage makes cross-view localization a highly scalable and practical solution for real-world deployment.

% Despite considerable progress, 
State-of-the-art cross-view localization methods primarily follow a fully supervised paradigm~\cite{xia2022visual, xia2023convolutional, wang2024fine, shi2023boosting, lentsch2023slicematch}, relying on precisely annotated ground camera poses for training. 
However, obtaining such annotations at scale is costly, often requiring fleets of mobile mapping vehicles equipped with expensive sensor kits to traverse the environment.
% the acquisition of which is prohibitively expensive and labor-intensive at scale. 
Furthermore, these methods often suffer significant performance degradation in the cross-area setting~\cite{xia2022visual, xia2023convolutional, wang2024fine, shi2023boosting, lentsch2023slicematch}, where test images originate from regions different from the training areas. 
This limitation significantly hinders the scalability of cross-view localization deployment.
% Moreover, fully supervised methods often exhibit a significant performance degradation when evaluated in cross-area settings ~\cite{xia2022visual, xia2023convolutional, wang2024fine, shi2023boosting, lentsch2023slicematch}, where the testing regions differ geographically from the training areas.
% This limitation highlights a critical challenge in achieving robust and generalizable cross-view localization.

\begin{figure}[t]
    \centering
    \includegraphics[width=\linewidth]{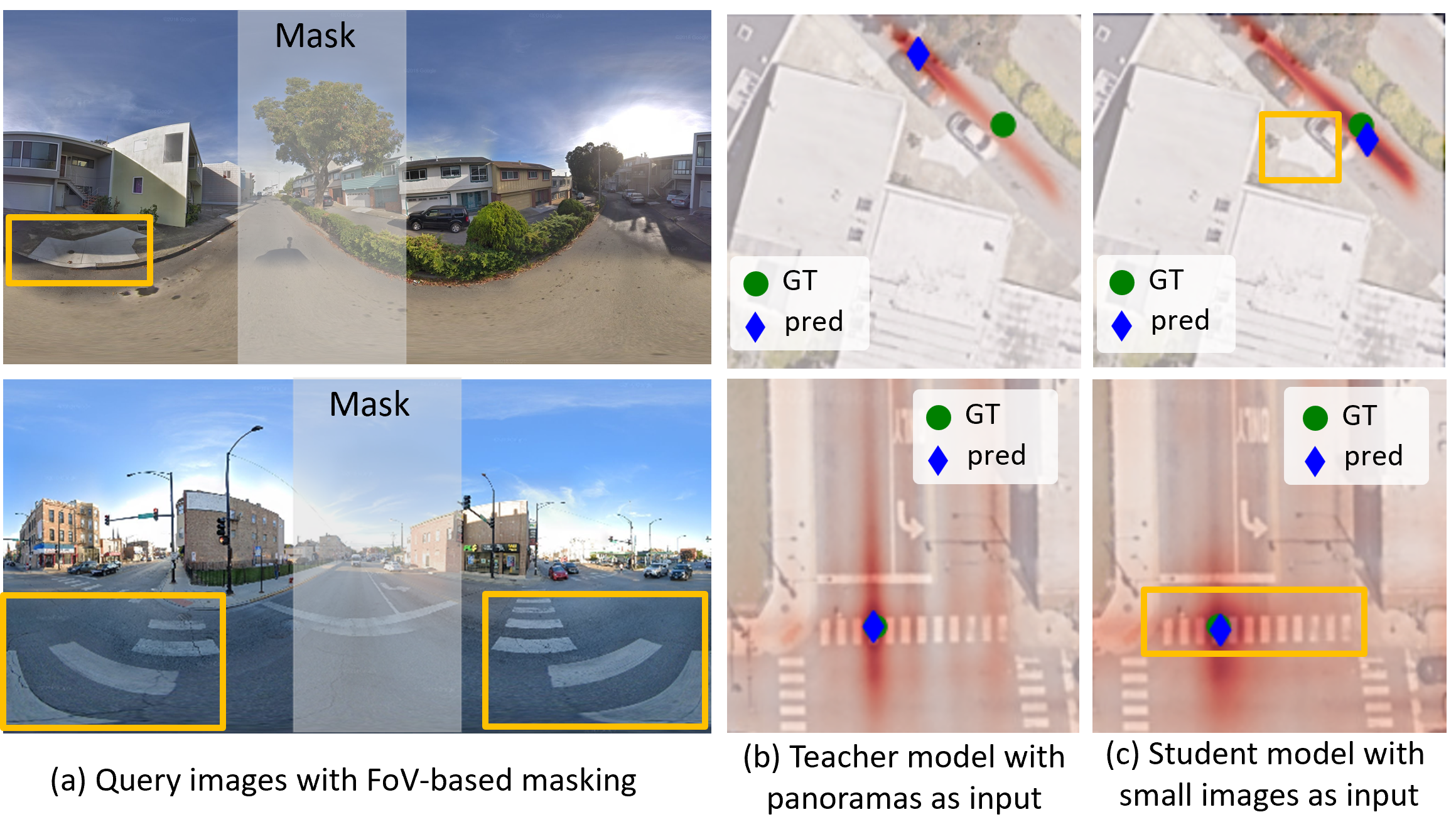}
    \caption{
    % We propose GeoDistill, a teacher-student knowledge self-distillation framework with FoV-based masking to enhance the learning of discriminative local features for weakly-supervised cross-view localization. 
    Our GeoDistill encourages the student model to extract discriminative local features from FoV-based masked inputs, resulting in more accurate localization (top right) and reduced uncertainty (bottom right). In contrast, the teacher model (middle), pre-trained on panoramas, lacks explicit enforcement for learning local features, leading to more uncertainty and wrong localization.
    % In our framework, the student model (right columns), which receives a query image with FoV-based masking (left columns), learns to focus on discriminative local features and predicts more accurate locations (right top) or shows reduced uncertainty (right bottom) compared to the teacher model (middle), which takes full panoramic images as input.
    }
    \label{fig:openfigure}
\end{figure}

To address this challenge, we propose leveraging weakly supervised learning.
While obtaining precise ground truth data is costly, collecting noisy ground truth remains accessible.
For example, images with coarse localization can be obtained using a mobile phone's GPS.
Using this coarse location, we can retrieve satellite images covering the local surroundings, allowing us to pair these images with the ground-level query image.
% We then utilize this coarse location to retrieve a satellite image covering the local surroundings and use the corresponding ground-satellite image pairs to enhance localization accuracy, particularly in the cross-area setting.
%

Recently, a few studies ~\cite{xia2024adapting, shi2024weakly} have explored this direction for 2-DoF location estimation with different formulations. 
\cite{xia2024adapting} assumes access to additional test-area data for fine-tuning, but in practice, collecting data from all test regions is infeasible. 
G2SWeakly~\cite{shi2024weakly} matches ground images with corresponding positive and negative satellite images, leveraging image-level deep metric learning objectives to achieve the goal of relative pose estimation between the ground and its positive satellite image. 
Although promising, image-level metric learning cannot provide strong supervision for accurate local features. These discriminative local features are essential to achieve high accuracy in estimating relative poses. 
% leverages ground-satellite image pairs for image-level metric learning. 
% Yet, we argue that scalable cross-view localization should rely on matching local features across views rather than solely learning discriminative image global descriptors.

This work aims to learn discriminative local features that generalize well in different areas using only ground-satellite image pairs. To do so, we propose a field-of-view (FoV)-based masking strategy for the query images to encourage the model to focus on different local features (See. Fig.~\ref{fig:openfigure} and Fig.~\ref{subfig:FoVMask} for some examples). Unlike patch- or pixel-based masking, which might remove critical information (such as ground structures) and retain irrelevant parts (like the sky), FoV-based masking ensures that the query image always includes a useful portion of the scene, enabling reasonable predictions. 
% This strategy also mimics the scenarios where a query image is captured by a camera with limited FoV.
%
A straightforward approach to leverage this FoV-based masking might be to use it as a data augmentation technique. 
% A straightforward approach is to mask out parts of the input image, using the automatic mask-out as a data augmentation strategy to encourage the model to focus on different local features. 
% encouraging the model to make correct predictions for both the original image and the image with masked-out regions, thus focusing on different local feature learning. 
However, this naive approach makes it harder for the model to learn generalizable features, as removing information from query images further complicates the already challenging cross-view localization task.
% fails to produce generalizable features, which will be demonstrated later, as removing information from query images further increases the difficulty of the already challenging cross-view localization task.
% thus impairing model learning. 

To tackle this challenge, we propose GeoDistill, a teacher-student knowledge self-distillation framework that explicitly encourages the model to make similar predictions for both panoramic images and masked images depicting the same scene.
% those with masked regions depicting the same scene. 
% , and design a teacher-student knowledge self-distillation~\cite{} pipeline to create a moving target for learning on masked images. 
Since panoramas typically provide more accurate location predictions than images with limited FoV, the teacher model takes the full panoramas as input, while the student model uses the FoV-based masked counterparts.
% with FoV-based masking. 
The student learns to mimic the teacher’s predictions, despite receiving less information.
% seeing a less rich scene. 
This forces the student to mine discriminative local features, such as lane markings, without relying on the overall global scene structure.
% focus on discriminative features, such as lane lines, 
% while ignoring textureless regions like roads.

As training progresses, the student model gradually learns to emphasize key local features and can outperform the teacher.
As shown in Fig.~\ref{fig:openfigure}, 
% In these two examples, 
the teacher model, which uses a panorama as input, makes incorrect predictions (top) or outputs high uncertainty along the road (bottom). 
In contrast, our student model, benefiting from enhanced local feature extraction, makes accurate predictions (top) and reduces uncertainty (bottom). 
% To further improve performance, 
Therefore, we update the teacher model’s weights by incorporating the student's weights through a moving average, progressively refining the teacher as a better learning target.
% using an exponential moving average of the student model’s weights after each training epoch. This iterative process ensures continuous improvement in the teacher model, raising the upper bound for the student’s performance.

Concretely, our contributions are summarized as follows:

\begin{itemize}
\item We introduce GeoDistill, a weakly supervised self-distillation paradigm that enhances local discriminative feature learning for robust cross-view localization. 
We demonstrate that this learning paradigm applies to different localization frameworks and improves their performance by over 10\% without architectural modifications.
% We demonstrate that this learning paradigm significantly improves the performance of different localization frameworks. 
% Its effectiveness is validated by retraining two state-of-the-art methods and achieving promising performance gains of 18.5\% and 13.6\% in more challenging cross-area settings without architectural modifications.

\item We explore FoV-based masking strategies to enhance localization-critical feature learning. We demonstrate that while FoV-based masking as a naive data augmentation impairs performance, it significantly improves the performance of localization frameworks when applied within our proposed teacher-student self-distillation pipeline.
% serves as a strong inductive bias in our self-distillation pipeline, guiding the network toward robust geometric matching.

\item Existing weakly supervised approaches lack the ability to estimate the relative orientation between ground-satellite images. We design an orientation estimation network that predicts the relative orientation between ground-satellite images without precise planar position ground truth.
\end{itemize}

\section{Related Work}
\label{sec:related}
\textbf{Cross-view localization} is typically formulated as either large-scale image retrieval~\cite{Liu_2019_CVPR, Regmi_2019_ICCV, shi2019spatial, yang2021cross, Zhu_2022_CVPR,Cai_2019_ICCV,mi2024congeo} or, more recently, fine-grained pose estimation. The latter, popularized by the VIGOR benchmark~\cite{zhu2021vigor}, aims to determine the precise 3-DoF pose between a ground image and its corresponding aerial view (e.g., satellite imagery~\cite{fervers2022uncertainty, shi2022accurate, shi2022beyond,xia2022visual, shi2023boosting, lentsch2023slicematch, xia2023convolutional, song2024learning, wang2024fine,xia2024adapting, shi2024weakly} or OpenStreetMap data~\cite{sarlin2023orienternet, sarlin2023snap}). 
The core challenge lies in the substantial visual disparity between viewpoints.
To bridge this viewpoint gap, one dominant strategy involves explicit geometric projections. Some approaches transform satellite imagery into a ground-view perspective using polar transformations~\cite{shi2022accurate, shi2022beyond}. Complementary methods project ground images into a bird's-eye-view (BEV) representation~\cite{fervers2022uncertainty, shi2023boosting, song2024learning, wang2024fine}. While these projection-based techniques have shown promise, they risk information loss during the transformation process. A contrasting line of work bypasses explicit geometry, instead employing end-to-end neural architectures that directly regress pose parameters from the original image pairs~\cite{xia2022visual, lentsch2023slicematch, xia2023convolutional}. This direct approach preserves all visual information but requires the model to implicitly learn more complex cross-view spatial relationships. 
While architecturally distinct, these approaches are universally designed for a fully supervised setting, demanding access to datasets with precise and costly ground-truth(GT) pose annotations.

More recently, weakly supervised paradigms have marked a significant advance~\cite{xia2024adapting, shi2024weakly}. These methods leverage noisy GPS data as a weak supervision signal, substantially reducing the dependency on precisely annotated GT. This innovation greatly enhances the scalability and deployability of cross-view localization in real-world scenarios, where obtaining accurate pose annotations can be prohibitively expensive or impractical.
However, these pioneering weakly supervised methods have their own limitations.
~\cite{xia2024adapting} simplifies the problem to 2-DoF translation by assuming a known camera orientation and requires in-domain data for adaptation. Meanwhile, while~\cite{shi2024weakly} successfully predicts orientation, its applicability is confined to ground image captured by standard pinhole cameras.

\textbf{Knowledge distillation}(KD)~\cite{buciluǎ2006model, gou2021knowledge, wang2021knowledge, 2021Emerging, allenzhu2023} aims to transfer knowledge from a more comprehensive teacher model to a more compact student model. Self-distillation(SD)~\cite{2020Self, zhang_selfdistill, Ji_2021_CVPR, huang_Comprehensiveself} stands as a compelling paradigm within KD frameworks. Unlike traditional KD, SD represents a specialized branch pioneered by Born-Again Networks~\cite{furlanello2018born}, wherein a student network learns from a teacher network with identical architecture, often initialized with the same weights. This approach elegantly circumvents the architectural constraints of conventional KD while preserving its knowledge transfer benefits.

SD effectively harnesses a network's inherent capacity by establishing a teacher-student dynamic through differentiated training procedures~\cite{zhang_be_2019, Kim_2021_ICCV} or strategically varied input data~\cite{xu_data-distortion_2019, Dong_2023_CVPR}. The fundamental insight of SD lies in its iterative refinement mechanism—using predictions from a previously trained model as target values for subsequent retraining cycles~\cite{2020Self}, thereby enabling the model to progressively distill and refine its own knowledge representation. This self-referential learning process has demonstrated remarkable efficacy in enhancing model performance without requiring additional architectural complexity or external knowledge sources.
\section{Methodology}
This section first formalizes the cross-view localization task and our weakly supervised learning setting.
Then, it introduces the details of our proposed GeoDistill method.

% This section defines the task of 3-DoF cross-view localization, which aims to estimate the pose between a ground-level image and a geo-referenced satellite image. We then introduce our proposed GeoDistill framework, which decomposes the task into rotation and translation predictions.

\subsection{Task Definition}
\label{sec:method}
 \begin{figure*}[h]
    \centering
    \includegraphics[width=\textwidth]{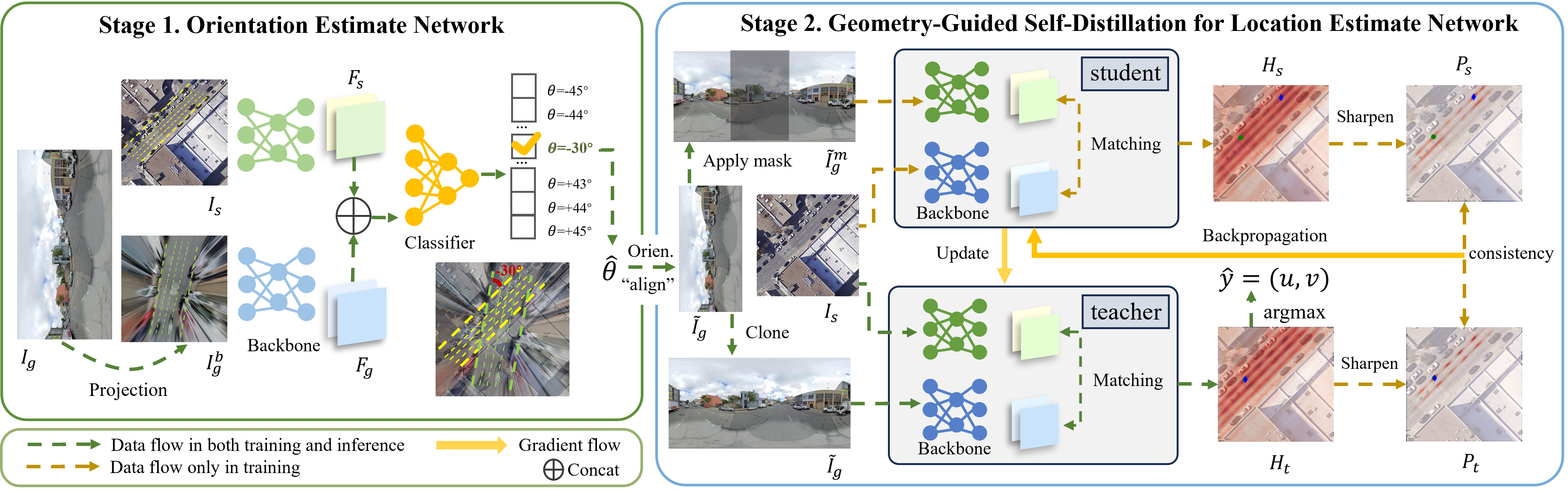}
    \caption{Overview of the proposed GeoDistill for 3-DoF ground-to-satellite relative pose estimation. Given a ground image, we first estimate its orientation with respect to the satellite image (Stage 1). For location estimation (Stage 2), we apply the proposed geometry-guided teacher-student self-distillation (GeoDistill) to a backbone framework, which can be any cross-view localization networks. All components in this pipeline are involved during training, while the green arrows indicate the workflow during inference. }
    % \caption{An overview of the proposed method for 3-DoF fine-grained cross-view localization. State 1 illustrates the 1-DoF orientation estimation network while state 2 demonstrates the 2-DoF location estimation network.}
    \label{fig: architecture}
\end{figure*}

Cross-view localization estimates the 3-DoF pose, i.e., the yaw $\theta$ and 2D translation $y = (u, v)$, between a ground-level image $I_g$ and a geo-referenced satellite image $I_s$. 
% Our task addresses cross-view camera localization by estimating the 3-DoF pose between a ground-level image $I_g$ and a geo-referenced satellite image $I_s$. 
% The pose consists of a yaw rotation $\theta$ and a 2D translation vector $y = (u, v)$. 
Current methods ~\cite{xia2023convolutional, shi2024weakly, xia2022visual, lentsch2023slicematch, shi2023boosting} train a neural network to generate a heat map $\mathbf{H}$ for localization, with the highest-confidence location serving as the predicted position,
\begin{equation}
    \hat{y} = \operatorname{argmax}_{u, v} \mathbf{H}(u, v).
\end{equation}

The neural network is typically trained in a supervised manner, relying on ground truth location and yaw for learning. However, acquiring precise ground truth location data is costly, whereas ground-level images with noisy GPS measurements are readily available\footnote{Obtaining accurate location data requires expensive mapping vehicles, while coarse location and yaw orientation can be easily obtained using built-in phone GPS and compasses.}.
Thus, weakly supervised learning presents a more practical alternative. 

Following~\cite{xia2024adapting, shi2024weakly}, the coarse location is used to identify a satellite image that covers the ground camera's location.
Our objective is then to use the ground-aerial image pairs to train a deep model for cross-view localization.

% leverages imprecise image pairs without requiring exact displacement annotations. Instead of relying on ground truth offsets, we use prior knowledge and visual correspondences to learn the correction function, enabling robust localization under realistic conditions.

% In practice, ground images often have noisy GPS measurements within a 40-meter radius of the true location. Given a ground image with noisy GPS $\hat{y}_g$ and a satellite image centered at this location, the true location is:
% \begin{equation}
%     y_g = \hat{y}_g + \Delta p \quad \text{where} \quad |\Delta p|\infty \leq 40m
% \end{equation}

% While these imprecise data are readily available, fully supervised methods require images $I_g$, $I_s$, and precise GT displacements $y = (u, v)$ during training. This demands accurate GPS coordinates for both images, making large-scale deployment expensive.

% Weakly supervised framework ~\cite{xia2024adapting, shi2024weakly} leverages imprecise image pairs without requiring exact displacement annotations. Instead of relying on ground truth offsets, we use prior knowledge and visual correspondences to learn the correction function, enabling robust localization under realistic conditions.

As shown in Fig.~\ref{fig: architecture}, we decompose the task into two sequential steps: 
% first, estimating the yaw orientation, followed by estimating the 2D translation.
% Since jointly predicting the full 3-DoF pose under weak supervision is challenging, we decompose the task into sequential 1-DoF rotation estimation and 2-DoF translation prediction, our framework illustrated in Fig.~\ref{fig: architecture}. 
First, our Rotation Estimator (Sec. \ref{sec: rotation network}) predicts the yaw angle $\hat{\theta}$, and uses it to rotate $I_g$ by horizontally shifting the panorama.
The transformed image $\widetilde{I_g}$ is then fed into a Location Estimator $f$ that predicts heat maps for localization
$\hat{\mathbf{H}} = f(\widetilde{I_g}, I_s)$.
Notably, our proposed Geometry-Guided Self-Distillation (Sec. \ref{sec: localization network}) is a generic weakly supervised learning paradigm compatible with various location estimators.

\subsection{Orientation Estimation}
\label{sec: rotation network}

% \begin{figure}[t!]
%     \centering
%     \includegraphics[width=0.45\textwidth]{geoDistill/figs/rotation.png}
%     \caption{Mean localization error of the student model when trained with different FoVs on VIGOR~\cite{zhu2021vigor} cross-Area test set.}
%     \label{fig: rotation}
% \end{figure}

The motivation of our orientation estimation network is to align prominent structural cues between ground and aerial views shared,  such as road layouts, which are typically the most dominant visual feature in outdoor scenes. However, a primary challenge arises from the inherent perspective distortion of panoramic images, which projects real-world straight roads into curves. To overcome this geometric mismatch, we employ a spherical transform~\cite{wang2024fine} to project $I_g$ into a Bird's-Eye-View (BEV) representation, $I_g^b$. This projection rectifies the road geometry, making it consistent with the top-down satellite image $I_s$ and thus enabling a direct comparison for orientation alignment.

Both the $I_s$ and $I_g^b$ are fed into a unshared feature extractor backbone, denoted as $\mathcal{E}_s(\cdot)$ and $\mathcal{E}_g(\cdot)$, respectively: 
\begin{equation}
    \mathbf{F}_s = \mathcal{E}_s(\mathbf{I}_s), \quad \mathbf{F}_g = \mathcal{E}_g(\mathbf{I}_g^b),
\end{equation}
where $\mathbf{F}_s, \mathbf{F}_g \in \mathbb{R}^{H \times W \times C}$. To facilitate cross-view feature interaction, we concatenate the extracted feature maps channel-wise to obtain a fused feature representation $\mathbf{F}$:
\begin{equation}
    \mathbf{F} = \text{Concat}(\mathbf{F}_s, \mathbf{F}_g), \quad \text{where }\mathbf{F} \in \mathbb{R}^{H \times W \times 2C}.
\end{equation}

The fused feature map $\mathbf{F}$ is processed by a Multilayer Perceptron (MLP) to predict orientation $\hat{\theta}$. To simplify the problem, we formulate this as a classification problem over discrete classes, one per degree in a predefined range. In the classification setup, the network outputs a probability distribution, and training uses the Cross-Entropy loss $L_{\text{CE}}$ between this distribution and the smoothed labels.
% Section~\ref{sec:Classification vs. Regression} show that both classification and regression approaches yield comparable performance.

\subsection{Geometry-Guided Self-Distillation}
\label{sec: localization network}
To achieve robust cross-view localization, a model must learn to identify discriminative local features. 
We propose a novel self-distillation method to directly cultivate this capability. 
At its core, the method enforces prediction consistency between a complete panorama and its partial views. 
Since the full and partial views depict the same geographical location, they must map to the identical satellite coordinate. 
This geometry-guided consistency requirement serves as a powerful supervision signal, compelling the model to discover salient local cues rather than depending on the brittle context of a full panorama.

% Our proposed Geometry-Guided Self-Distillation (GeoDistill) aims to improve local feature learning for robust cross-view localization.
% % In this section, we present GeoDistill, a weakly supervised knowledge self-distillation framework designed to improve local feature learning for robust cross-view localization. 
% % GeoDistill 
% It leverages a teacher-student knowledge distillation paradigm with asymmetric inputs and an uncertainty preservation strategy. 
% Next, we provide details for the proposed method. 

% Next, we describe its key components, including the teacher-student setup, the FoV-based masking strategy, and the training procedure.

% While 360° panoramas typically outperform limited FoV inputs, some limited-FoV cases achieve better accuracy (Fig.~\ref{fig:openfigure}). This indicates reliance on spurious correlations rather than true geometric relationships, as proper geometric matching would not yield limited views outperform complete ones. Grounded on this, we propose geometry guided self distillation paradigm to refine the model by aligning full and limited-view outputs, with an uncertainty preservation strategy to handle cases where improper matching initially causes student predictions to surpass the teacher.

The teacher model is a pre-trained Location Estimator $f_t(\cdot; \theta_t)$, where $\theta_t$ denotes its weights. 
It can be any recent cross-view localization method that generates heat maps for localization~\cite{xia2023convolutional, shi2024weakly, xia2022visual, lentsch2023slicematch, shi2023boosting}.
The student model $f_s(\cdot; \theta_s)$ has the same architecture as the teacher and is initialized with the teacher's weights, i.e., $\theta_s=\theta_t$ at the start of training.
A key design in GeoDistill is that the teacher and student will receive different inputs, forcing them to extract distinct features for localization.

% We implement this by duplicating a pre-trained model into teacher and student networks with identical architectures and initial weights, denoted as $f_t(\cdot; \theta_t)$ and $f_s(\cdot; \theta_s)$. The only difference lies in their input information richness, creating a performance gap where the teacher provides stable learning targets while the student explores proper geometric relationships to improve itself.
\textbf{FoV-based geometric consistency.} 
Our geometric guidance enforces prediction consistency between a full view and a partial view of the same scene, which we achieve via an teacher-student architecture. 
The teacher model $f_t$ receives the transformed 
full $360^\circ$ panorama $\widetilde{I_g}$ and the satellite image $I_s$.
This is identical to standard cross-view localization methods~\cite{xia2023convolutional, shi2024weakly, xia2022visual, lentsch2023slicematch, shi2023boosting}, where models typically construct a complete feature map of the scene~\cite{shi2024weakly} or learn global image descriptors~\cite{xia2023convolutional} for localization.

To encourage the student $f_s$ to explore local features in the image, we do not feed the full panorama to it.
Instead, we apply a mask with random FoV to the panorama $\widetilde{I_g}$,
% to simulate a limited FoV:
\begin{equation}
        \widetilde{I_g}^m = M(\widetilde{I_g}), \quad M(x) = x \odot M_{\text{mask}},
\end{equation}
where $M_{\text{mask}} \in \{0,1\}^{H \times W}$.
The resulting masked image $\widetilde{I_g}^m$ simulates an image with a limited FoV.
Fig.~\ref{subfig:FoVMask} illustrates two examples of the FoV-based masking. 

During training, the teacher network's weights, $\theta_t$, are frozen, serving as a provider of stable and reliable learning targets. Both teacher and student networks process their respective inputs, along with the satellite image ${I_s}$. to generate heat maps:
\begin{equation}
        H_t = f_t(\widetilde{I_g}, I_s; \theta_t), \quad H_s = f_s(\widetilde{I_g}^m, I_s; \theta_s).
\end{equation}

This design creates a geometrically-grounded learning task where the student must learn to produce a localization output consistent with the teacher's, but from incomplete information. This forces the student to move beyond reliance on the global scene structure and instead mine for robust, discriminative local features for accurate localization.

\textbf{Uncertainty preservation.} 
Although the teacher model provides a learning target, its predictions inevitably contain noise. 
The student's predictions are also noisy, exacerbated by its less informative input. Directly aligning these noisy distributions is difficult. While one solution might be to use a hard, one-hot target, this would discard the valuable dark knowledge encoded in the teacher's heatmap. This knowledge, encoded in the distribution's shape and relative activation strengths, implicitly teaches the student about structural similarities and model uncertainty.

We adopt a more nuanced strategy: sharpening both heatmaps before calculating the distillation loss. 
This serves a dual purpose: it reduces the distribution's entropy, compelling the student to focus on the teacher's highest-confidence predictions, while simultaneously mitigating the impact of noisy, low-confidence signals. 
Specifically, we apply a softmax with a temperature $\tau<1$ to both outputs: 
\begin{equation} 
P_t = \text{Softmax}(H_t / \tau), \quad P_s = \text{Softmax}(H_s / \tau). 
\end{equation}

\textbf{Bidirectional knowledge flow.} The student learns from the teacher by minimizing the self-distillation loss \( \mathcal{L}_{SD} \), defined as the Cross-Entropy between their sharpened output distributions \( P_s \) and \( P_t \): 
\begin{equation} 
\mathcal{L}_{SD} = \mathbb{E}{X} \left[ - \sum_{i} P_t(i) \log P_s(i) \right]. 
\end{equation} 
While the student's weights \( \theta_s \) are updated via gradient descent, we facilitate a bidirectional knowledge flow~\cite{2020Momentum, 2021Emerging} by updating the teacher's weights \( \theta_t \) as an Exponential Moving Average (EMA) of the student's:
\begin{equation} 
\theta_t \leftarrow \alpha \theta_t + (1-\alpha) \theta_s . 
\end{equation} 
This EMA mechanism allows the teacher to progressively absorb the robust features learned by the student. For inference, we use this continuously refined teacher model to achieve enhanced performance.

\section{Experiments}
\label{sec:main results}

% \begin{figure*}
%     \centering
%     \begin{subfigure}{0.33\linewidth}
%         \centering
%         \includegraphics[width=\linewidth]{geoDistill/figs/d1.png}
%     \end{subfigure}
%     \begin{subfigure}{0.33\linewidth}
%         \centering
%         \includegraphics[width=\linewidth]{geoDistill/figs/d2.png}
%     \end{subfigure}
%     \begin{subfigure}{0.33\linewidth}
%         \centering
%         \includegraphics[width=\linewidth]{geoDistill/figs/d3.png}
%     \end{subfigure}
%     \caption{Qualitative comparison of probability maps before (left) and after (right) applying GeoDistill. The first row presents input panoramic scenes, while the second row shows the predicted localization heat maps. Red indicates localization probability, with darker shades representing higher probabilities.}
%     \label{fig: before vs after}
% \end{figure*}

 \begin{figure*}[h]
    \centering
    \includegraphics[width=\textwidth]{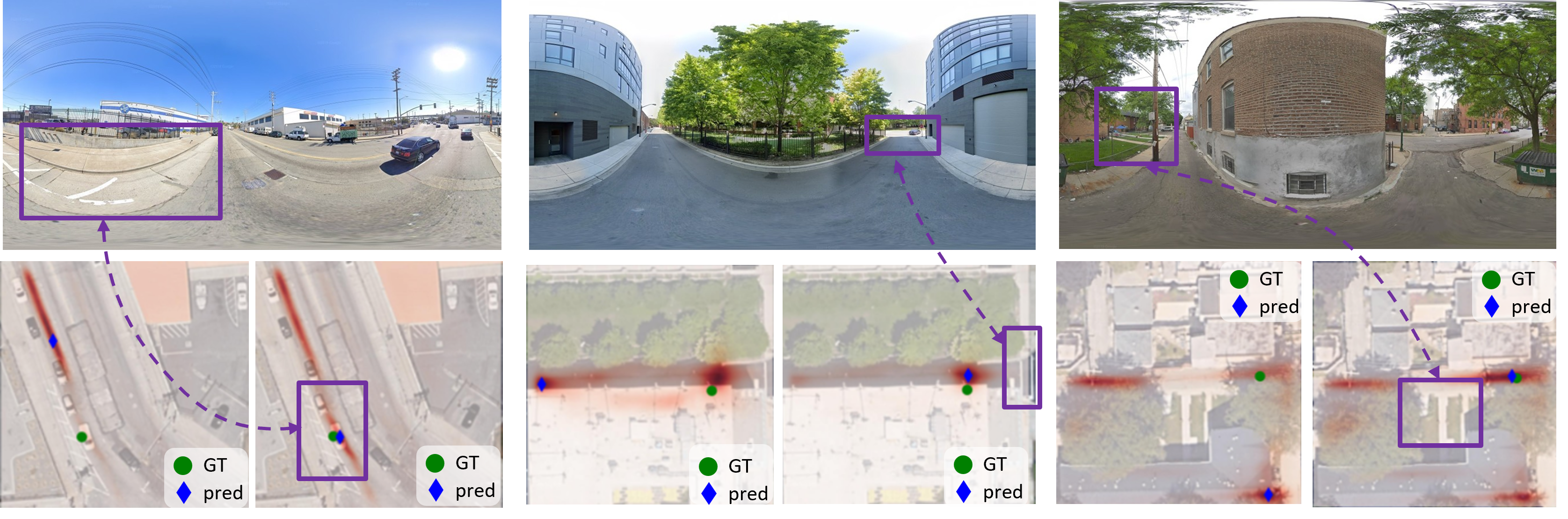}
    \caption{Qualitative comparison of probability maps before (left) and after (right) applying proposed self-distillation learning paradigm on VIGOR~\cite{zhu2021vigor} Cross-Area test set. The first row presents input panoramic scenes, while the second row shows the predicted localization heat maps. Red indicates localization probability, with darker shades representing higher probabilities.}
    \label{fig: before vs after}
\end{figure*}

%  \begin{figure*}[h]
%     \centering
%     \includegraphics[width=\textwidth]{geoDistill/figs/beforeVSafter.pdf}
%     \caption{Qualitative comparison of probability maps before (left) and after (right) applying GeoDistill. The first row presents input panoramic scenes, while the second row shows the predicted localization heat maps. Red indicates localization probability, with darker shades representing higher probabilities.}

%     \label{fig: before vs after}
% \end{figure*}

In this section, we conduct experiments to answer the following questions: 
(1) 
% What is the necessity and effectiveness of each component in the proposed GeoDistill approach?
How necessary and effective is each component of the proposed GeoDistill approach? (Sec.~\ref{subsec:model_analysis}). 
(2) 
% What is its performance over other weakly supervised approaches? Will the proposed GeoDistill learning paradigm also improve other frameworks' performance when trained with full supervision? 
How does GeoDistill compare to other weakly supervised methods? Can it also enhance fully supervised methods?
(Sec.\ref{subsec:sota}).
(3) 
% How is its performance gap with fully supervised state-of-the-art? 
What is the performance gap between GeoDistill and fully supervised state-of-the-art methods?
(Sec.~\ref{subsec:fully}).
% This section first introduce the dataset and evaluation metrics, followed by the analysis of GeoDistill's core components. We then discuss two state-of-the-art methods \cite{xia2023convolutional, shi2024weakly} serving as baselines for our weakly-supervised evaluation.  Finally, we present comprehensive performance improvements achieved with GeoDistill.
First, we outline the dataset, evaluation metrics, base models, and implementation details.

% \subsection{Datasets and Evaluation Metrics}
\subsection{Datasets and Evaluation Metrics}
 We evaluate GeoDistill on two different datasets:
\begin{itemize}
    \item \textbf{VIGOR} 
    % comprise 105,214 geo-referenced ground-level panoramas and corresponding aerial imagery from four US cities. Satellite images in VIGOR~\cite{zhu2021vigor} correspond to a ground area of approximately 70m × 70m. Following the dataset protocol, an aerial patch is labeled as positive if its central quarter encompasses the ground camera location; otherwise, it is deemed semi-positive. Consistent with prior work, we utilize only positive aerial patches for both training and evaluation across all experiments. The VIGOR~\cite{zhu2021vigor} dataset provides an orientation prior, with panoramas pre-aligned to North. During training, orientation labels are generated by perturbing this prior with ±45° noise, simulating realistic sensor inaccuracies and augmenting rotational variance.
    % We adopt the Same-Area and Cross-Area data splits as defined by~\cite{zhu2021vigor}.  For validation and hyperparameter optimization, 20\% of the training set was randomly held out, following established practices~\cite{xia2023convolutional, shi2024weakly, wang2024fine, shi2023boosting}. 
    contains 105,214 pairs of geo-referenced ground panoramas and corresponding aerial images from four US cities, with each aerial image covering a 70m × 70m area. Following the official protocol, we only use positive pairs—where the ground camera's location is within the central quarter of the aerial image—for all training and evaluation. Panoramas are North-aligned in VIGOR~\cite{zhu2021vigor}, we augment the panoramas by applying random orientation noise within a ±45° range to generate orientation label. We adopt the standard Same-Area and Cross-Area splits for evaluation. For hyperparameter tuning, a validation set is created by holding out 20\% of the training data, consistent with prior works\cite{xia2023convolutional, shi2024weakly, wang2024fine, shi2023boosting}.
    
    \item \textbf{KITTI}~\cite{geiger2013vision} provides limited-FoV ground images captured by pin-hole camera from Germany, coupled with aerial views from~\cite{shi2022accurate}. Following the standard setup~\cite{shi2022beyond}, we use the same-area and cross-area splits, where ground camera locations are within a central 40m × 40m aerial patch and an orientation prior with ±10° noise is given.
\end{itemize}

\textbf{Evaluation metrics}. Performance evaluation is conducted using standard metrics: mean and median errors, computed separately for localization (in meters) and orientation (in degrees), across all test samples, providing a comprehensive assessment of accuracy.

\textbf{Base models}. 
% We validate the effectiveness of the proposed GeoDistill based on G2SWeakly~\cite{shi2024weakly}, a weakly supervised approach that does not require GT data for model training, ensuring that the entire pipeline is strictly weakly supervised. 
% Furthermore, we also demonstrate that the proposed GeoDistill is able to improve the performance of a fully supervised method without using GT labels. 
% For this, we employ CCVPE~\cite{xia2023convolutional} as our additional backbone. 
% Both G2Sweakly~\cite{shi2024weakly} and CCVPE~\cite{xia2023convolutional} solve the problem of fine-grained cross-view localization.
% CCVPE~\cite{xia2023convolutional} employs descriptor-based matching and trains its network using vanilla panoramas, while G2SWeakly~\cite{shi2024weakly} projects panoramas to generate BEV images to mitigate the visual discrepancy between ground and satellite views.
We validate the effectiveness and broad applicability of the proposed GeoDistill on two distinct fine-grained cross-view localization methods: G2SWeakly~\cite{shi2024weakly} and CCVPE~\cite{xia2023convolutional}.
Our primary experiments build upon G2SWeakly, a state-of-the-art weakly supervised approach that projects panoramas into BEV images to mitigate visual discrepancies. 
\textit{This ensures our entire pipeline remains strictly weakly supervised}. 
While the original G2SWeakly uses a VGG backbone, we also implement a variant using DINOv2 to leverage its powerful, generalizable feature representations.
Furthermore, to demonstrate the versatility of GeoDistill, we apply it to CCVPE~\cite{xia2023convolutional}, a fully supervised method that relies on descriptor-based matching and trains its network using vanilla panoramas.

\textbf{Implementation details.} 
% For localization estimation network, we evaluate our weakly supervised self distillation paradigm to different state-of-the-art methods, CCVPE ~\cite{xia2023convolutional} and G2SWeakly ~\cite{shi2024weakly}.
For location estimation, we use the code released by the authors of CCVPE~\cite{xia2023convolutional} and G2SWeakly~\cite{shi2024weakly} for model implementations. Following the two model’s default settings, we use a batch size of 8, and a learning rate of 0.0001 with Adam optimizer~\cite{2014Adam} for both models. The temperature $\tau$ is set to 0.06 for both teacher and student. The EMA ritio $\alpha$ is 0.9. For G2SWeakly~\cite{shi2024weakly}-DINO variant, we employ a pre-trained DINOv2-b14 as the feature extractor. During training, the weights of the DINOv2 are kept frozen, and we append a DPT~\cite{Ranftl2021Vision} module to fine tune DINO feature.
Our orientation estimation framework employs EfficientNet-B0~\cite{2019EfficientNet} with pretrained weights on Imagenet~\cite{deng_imagenet} as both the ground and aerial feature extractors, with non-shared weight. The satellite image and BEV transformed from the ground image both have a size of 512 × 512. All experiments were conducted using a single NVIDIA 4090 GPU.

% \subsection{Performance Improvements via GeoDistill}
% \subsection{Comparison with the State of the Art}
\subsection{Effectiveness of Proposed Distillation Paradigm}
\label{subsec:sota}

\begin{table*}
\centering
\begin{tblr}{
  width = \linewidth,
  % 第二列默认左对齐 l
  colspec = {c | l | X[c] X[c] | X[c] X[c]}, 
  rowsep = 1pt,
  column{3-6} = {c},
  cell{1}{1} = {r=2}{},
  % 1. 这里只负责结构（跨行），不负责格式
  cell{1}{2} = {r=2}{},
  cell{1}{3} = {c=2}{},
  cell{1}{5} = {c=2}{},
  cell{3,9}{1} = {r=6}{},
  hline{1,3,9,15} = {-}{},
  hline{5,11} = {2-6}{},
  hline{7,13} = {2-6}{dashed},
}
% --- 表格内容区 ---
% 2. 使用 \SetCell 命令对特定单元格进行居中设置
Dataset & \SetCell{halign=c} Method            & Cross-Area                         &                                    & Same-Area                          &                                    \\
        &                                      & $\downarrow$Mean(m)                & $\downarrow$Median(m)              & $\downarrow$Mean(m)                & $\downarrow$Median(m)              \\
VIGOR   & CCVPE~\cite{xia2023convolutional}    & 4.97                               & 1.68                               & 3.60                               & 1.36                               \\
        & + \textbf{GeoDistill}                & \textbf{4.05} ($\downarrow$18.5\%) & \textbf{1.57} ($\downarrow$6.5\%)  & \textbf{3.21} ($\downarrow$10.8\%) & \textbf{1.31} ($\downarrow$3.7\%)  \\
        & G2SWeakly~\cite{shi2024weakly}(VGG)  & 5.20                               & 1.44                               & 4.81                               & 1.61                               \\
        & + \textbf{GeoDistill}(VGG)           & \textbf{4.49} ($\downarrow$13.6\%) & \textbf{1.22} ($\downarrow$15.3\%) & \textbf{4.26} ($\downarrow$11.4\%) & \textbf{1.37} ($\downarrow$14.9\%) \\
        & G2SWeakly~\cite{shi2024weakly}(DINO) & 3.58                               & 1.45                               & 3.61                               & 1.59                               \\
        & + \textbf{GeoDistill}(DINO)          & \textbf{2.68} ($\downarrow$25.1\%) & \textbf{1.20} ($\downarrow$17.2\%) & \textbf{3.08} ($\downarrow$14.7\%) & \textbf{1.39} ($\downarrow$12.6\%) \\
        \hline
KITTI   & CCVPE~\cite{xia2023convolutional}    & 8.94                               & 3.33                               & 1.28                               & 0.71                               \\
        & + \textbf{GeoDistill}                & \textbf{6.99}($\downarrow$21.8\%) & \textbf{3.14} ($\downarrow$5.7\%)  & \textbf{1.25} ($\downarrow$0.2\%) & \textbf{0.71} ($\downarrow$0\%)                      \\
        & G2SWeakly~\cite{shi2024weakly}(VGG)  & 12.54                              & 10.56                              & 11.11                              & 9.74                               \\
        & + \textbf{GeoDistill}(VGG)           & \textbf{12.16} ($\downarrow$3.0\%) & \textbf{10.22} ($\downarrow$3.2\%) & \textbf{10.97} ($\downarrow$1.3\%) & \textbf{9.62} ($\downarrow$1.2\%)\\
        & G2SWeakly~\cite{shi2024weakly}(DINO) & 12.61                              & 11.64                              & 11.68                              & 10.96                              \\
        & + \textbf{GeoDistill}(DINO)          & \textbf{11.85} ($\downarrow$6.0\%) & \textbf{11.17} ($\downarrow$4.0\%) & \textbf{11.52} ($\downarrow$1.4\%) & \textbf{10.91} ($\downarrow$0\%)                     
\end{tblr}
\caption{Localization performance improvement over different baselines on VIGOR~\cite{zhu2021vigor} and KITTI test set. Our proposed self distillation learning paradigm consistently improves base models' performance without access to ground truth location labels.}
\label{tab: diff baseline}
\end{table*}

To validate the effectiveness and generalizability of GeoDistill, we conduct a comprehensive evaluation. First, we assess the proposed self-distillation paradigm by incorporating it into two distinct base models: the weakly supervised G2SWeakly~\cite{shi2024weakly} and the fully supervised CCVPE~\cite{xia2023convolutional}. Subsequently, we combine the orientation and location estimators to evaluate the full 3-DoF pose estimation performance, with a detailed comparison against fully supervised methods presented in Sec.~\ref{subsec:fully}. Experiments are conducted on the VIGOR and KITTI datasets. Notably, a key advantage of our method is its minimal supervision requirement: it operates solely on orientation-aligned ground-satellite image pairs, without the need for GT translation. This holds even when GeoDistill is applied to CCVPE, a model that conventionally requires complete GT poses for its training.

\textbf{Results on VIGOR}. 
Tab.~\ref{tab: diff baseline}(top half) highlights the significant performance gained by GeoDistill on the VIGOR dataset. 
For the CCVPE~\cite{xia2023convolutional}, which suffers from a notable performance disparity between evaluation settings, our distillation proves highly effective. 
It yields the most pronounced gains in the challenging cross-area scenario, effectively narrowing the performance gap and enhancing the model's generalization.
When combined with the G2SWeakly~\cite{shi2024weakly}, it consistently yields substantial improvements. Notably, the performance lift is markedly greater for the stronger DINO-based variant. 
This suggests our method's efficacy scales with the base model's quality, as a superior feature extractor like DINO provides a richer foundation from which to distill nuanced knowledge, leading to more pronounced gains.

% Tab.~\ref{tab: diff baseline vigor} highlights the significant performance gained by GeoDistill on the VIGOR dataset. For the G2SWeakly backbone, which is already weakly-supervised, our method yields a substantial overall performance improvement of over 10\% in both same-area and cross-area evaluations. For the CCVPE backbone, we observe that the original model suffers from a significant performance disparity between same-area and cross-area settings. Notably, GeoDistill provides a more pronounced performance gain in the challenging cross-area scenario, effectively narrowing this performance gap.

\textbf{Results on KITTI}. We further test on the KITTI dataset, which presents a new challenge with its limited FOV captured by pinhole images, unlike VIGOR's $360^\circ$ panoramas.
Tab.~\ref{tab: diff baseline}(bottom half) shows that results on KITTI follow a similar trend. After distillation, CCVPE delivers substantial gains in the cross-area split. In the same-area setting, however, performance is unchanged, as the baseline's already low localization error creates a performance ceiling.
When applied to G2SWeakly, our paradigm improves both VGG and DINO variants, though the gains are more modest than on VIGOR. We attribute this to the baseline's weaker initial performance on KITTI, which provides a noisier supervisory signal for distillation.

Overall, the consistent performance gains across different models and datasets validate our self-distillation paradigm as an effective, plug-and-play solution for self-improvement without architectural modifications.

\textbf{Qualitative results.}
Fig.~\ref{fig: before vs after} qualitatively compares the location probability maps predicted by the G2SWeakly~\cite{shi2024weakly} as base model on the VIGOR dataset, both with and without our distillation paradigm. As highlighted by purple rectangles, the vanilla G2SWeakly~\cite{shi2024weakly} struggles when localization should depend on local features. It often assigns high probability to incorrect locations while attributing low probability to the GT location, indicating poor cross-view feature matching. In contrast, our distillation-enhanced model effectively leverages these discriminative local features to produce accurate and confident location predictions. This demonstrates our method's ability to significantly improve feature representation for cross-view localization.

\subsection{Comparison with Fully Supervised Methods}
\label{subsec:fully}
We assess the generalization capabilities about 3-DoF pose estimation of GeoDistill by conducting a rigorous evaluation on the VIGOR cross-area setting. We compare our method against several fully supervised approaches. Among them, HC-Net~\cite{wang2024fine} represents the current state-of-the-art performance. We evaluate performance under varying levels of orientation prior noise, including $0^\circ$ and $\pm 45^\circ$ for ground images. \textit{Notably, we implement GeoDistill based on G2SWeakly~\cite{shi2024weakly} to ensure entire pipeline remains strictly weakly supervised, requiring no GT location labels.}

Tab.~\ref{tab: cross area performance} shows GeoDistill's effectiveness. With a VGG backbone, GeoDistill already achieves the best median localization error among all methods, surpassing the fully supervised SOTA. Although its mean error is marginally higher in this setup, its superior median performance highlights its robustness. 
This advantage becomes definitive when using a DINO backbone. Our weakly supervised model then surpasses all fully supervised methods across both mean and median metrics. 
% This progression demonstrates that GeoDistill effectively leverages advanced feature extractors, enabling a weakly supervised system to outperform fully supervised state-of-the-art.
Additionally, GeoDistill surpasses all competing approaches in orientation estimation, including those that are fully supervised, demonstrating the effectiveness of our orientation estimation network.

\begin{table}[htbp] % 使用 [htbp] 让 LaTeX 自动选择最佳位置
\centering % 使用 \centering 代替 center 环境，更标准
\begin{tblr}{
  % 使用 width=\linewidth 让表格自动适应行宽，不再需要 resizebox
  width = \linewidth,
  % 使用 X 列类型自动调整列宽，c=居中, l=左对齐, m=垂直居中
  colspec = {Q[c,m] Q[l,m] *{4}{X[c,m]}},
  % --- 表格线设置 ---
  % 统一管理所有横线，代码更清晰
  % hline{1} 是顶线, hline{3} 是表头下的线, hline{9} 是 0° 数据块内的分隔线, 等等
  hline{1, 3, 9, 11, 13, 15} = {-}{}, 
  % 统一管理所有竖线
  vline{2, 3, 5} = {-}{},
  % 设置特定行的格式，这里表头两行都设为居中
  row{1,2} = {c, m},
  rows = {rowsep=0pt},
}
% --- 表头 ---
% 使用 \SetCell 进行列合并，取代了 \multicolumn
\SetCell[r=2]{c} Noise & \SetCell[r=2]{c} Method & \SetCell[c=2]{c} $\downarrow$Localization & & \SetCell[c=2]{c} $\downarrow$Orientation & \\
      &        & Mean & Median & Mean & Median \\
% --- 数据部分 ---
% 第一组数据：0°
\SetCell[r=8]{c} 0° & CVR~\cite{zhu2021vigor}*          & 9.45          & 8.33          & -             & -             \\
                     & SliceMatch~\cite{lentsch2023slicematch}* & 5.53          & 2.55          & -             & -             \\
                     & Boosting~\cite{shi2023boosting}*       & 5.16          & 1.40          & -             & -             \\
                     & CCVPE~\cite{xia2023convolutional}*      & 4.97          & 1.68          & -             & -             \\
                     & DenseFlow~\cite{song2024learning}*     & 5.01          & 2.42          & -             & -             \\
                     & HC-Net~\cite{wang2024fine}*            & 3.35          & 1.59          & -             & -             \\
                     & \textbf{Ours}(VGG)                     & 4.49          & 1.22          & -             & -             \\
                     & \textbf{Ours}(DINO)                    & \textbf{2.68} & \textbf{1.20} & -             & -             \\
% 第二组数据：±45°
\SetCell[r=4]{c} ±45° & CCVPE~\cite{xia2023convolutional}* & 5.16          & 1.78          & 26.77         & 15.29         \\
                      & HC-Net~\cite{wang2024fine}*       & \textbf{3.46} & 1.60 & 3.00 & \textbf{1.35} \\
                      & \textbf{Ours}(VGG)                & 4.99          & \textbf{1.33} & \textbf{2.72} & \textbf{1.35} \\ % 修正：将下划线改为了4.99，因为原代码如此
                      & \textbf{Ours}(DINO)               & 4.20          & 2.57 & \textbf{2.72} & \textbf{1.35} \\ % 修正同上
\end{tblr}
\caption{
Performance comparison with fully supervised approaches on 2-DoF and 3-DoF pose estimation on VIGOR~\cite{zhu2021vigor} Cross-Area test set. Here, GeoDistill employs G2SWeakly as base model, ensuring the model is trained strictly with weak supervision. ``*'' indicates fully supervised methods. \textbf{Best in bold}.
}
\label{tab: cross area performance}
\end{table}

\subsection{Model Analysis}
\label{subsec:model_analysis}
% In this section, we conduct experiments to evaluate the core components of our GeoDistill paradigms, including the FoV-based masking, the uncertainty preservation strategy, and the teacher-student weight update strategy. 

% To validate the core components of GeoDistill, we adopt the weakly supervised state-of-the-art, G2SWeakly~\cite{shi2024weakly}, as a baseline, to perform a series of ablation studies, including the FoV-based masking, the uncertainty preservation strategy, and the teacher-student parameter update strategy.  
To validate the core components of GeoDistill, we conduct a series of ablation studies. 
We build upon the state-of-the-art weakly supervised method, G2SWeakly~\cite{shi2024weakly}, as our base model. 
\textit{Unless stated otherwise, all experiments utilize its original implementation with a VGG backbone. }
Our analysis investigates the contributions of our key components: the FoV-based masking, the uncertainty preservation strategy, and the teacher-student parameter update strategy.

\begin{table}
\centering
\begin{tblr}{
    colspec = {l l c c},
    column{3,4} = {c},
    cell{2}{1} = {r=4}{c},
    cell{6}{1} = {r=3}{c},
    vline{2} = {1-8}{},
    hline{1-2,6,9} = {-}{},
    hline{5,8} = {2-4}{},
    colsep = 6pt,
    rows = {rowsep=0pt},
}
Backbone & Mask & $\downarrow$Mean(m) & $\downarrow$Median(m) \\
CNN & Maximum Act. & 5.14 & 1.42 \\
& Random Patch & 5.21 & 1.44 \\
& FoV & \textbf{4.49} & \textbf{1.22} \\
& Baseline & 5.20 & 1.44 \\
\hline
ViT & Random Patch & 3.10 & 1.33 \\
& FoV & \textbf{2.68} & \textbf{1.20} \\
& Baseline & 3.58 & 1.45 \\
\end{tblr}
\caption{Performance comparison with different masking strategies on VIGOR~\cite{zhu2021vigor} Cross-Area test set.}
\label{tab: diff mask}
\end{table}
\begin{figure}[t]
    \centering
    \setlength{\abovecaptionskip}{0pt}
\setlength{\belowcaptionskip}{0pt}
    \begin{subfigure}[b]{0.48\linewidth}
        \includegraphics[width=\linewidth]{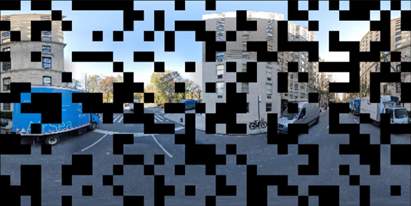}
        \caption{Random patch masking}
    \end{subfigure}
    \begin{subfigure}[b]{0.48\linewidth}
        \includegraphics[width=\linewidth]{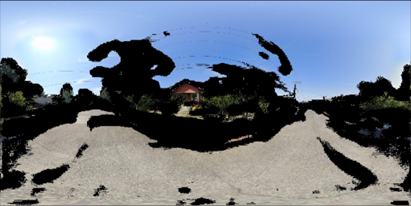}
        \caption{Maximum activation masking}
    \end{subfigure}
    \begin{subfigure}{0.98\linewidth}
        \includegraphics[width=0.49\linewidth]{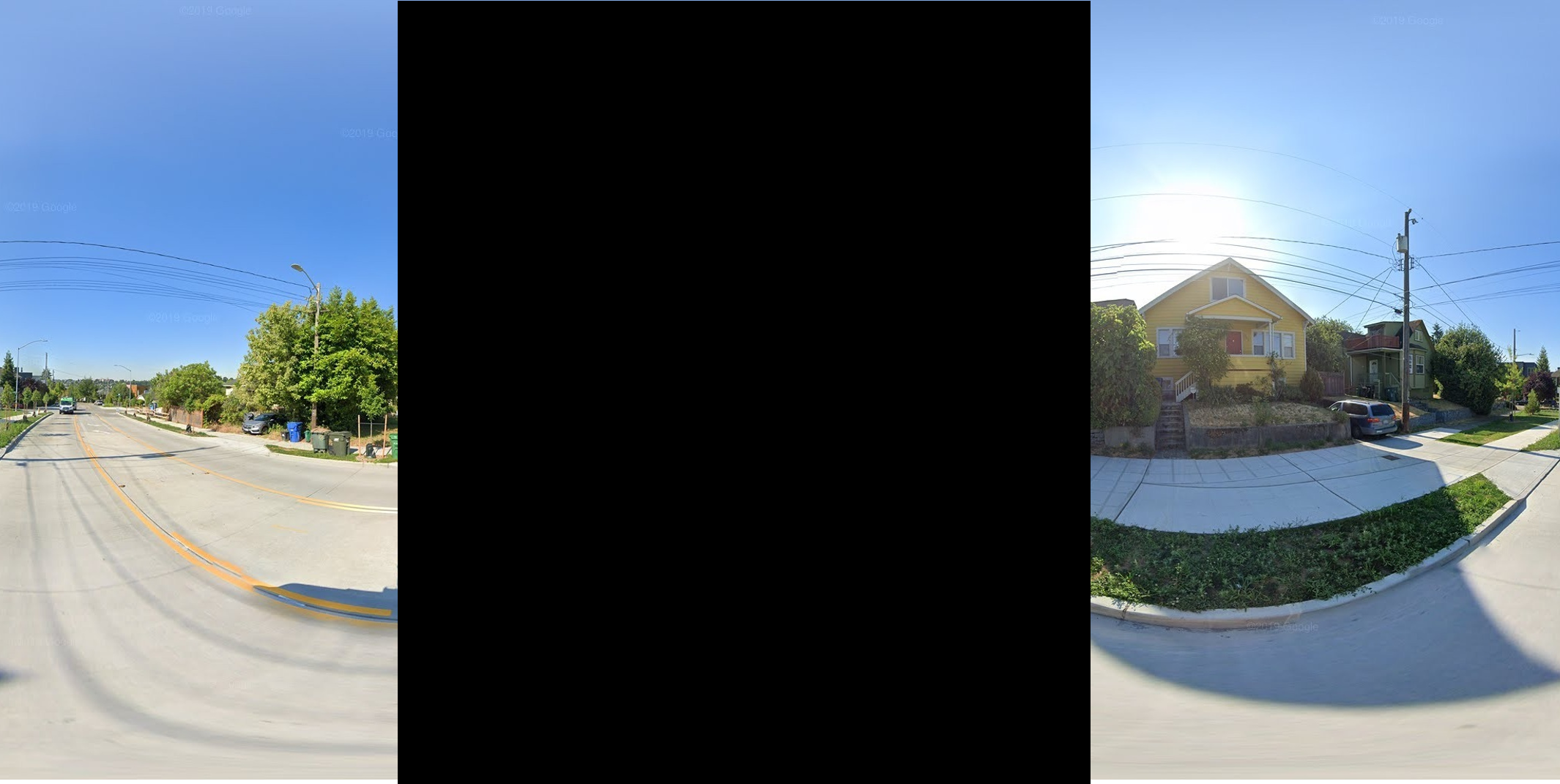}
        \includegraphics[width=0.49\linewidth]{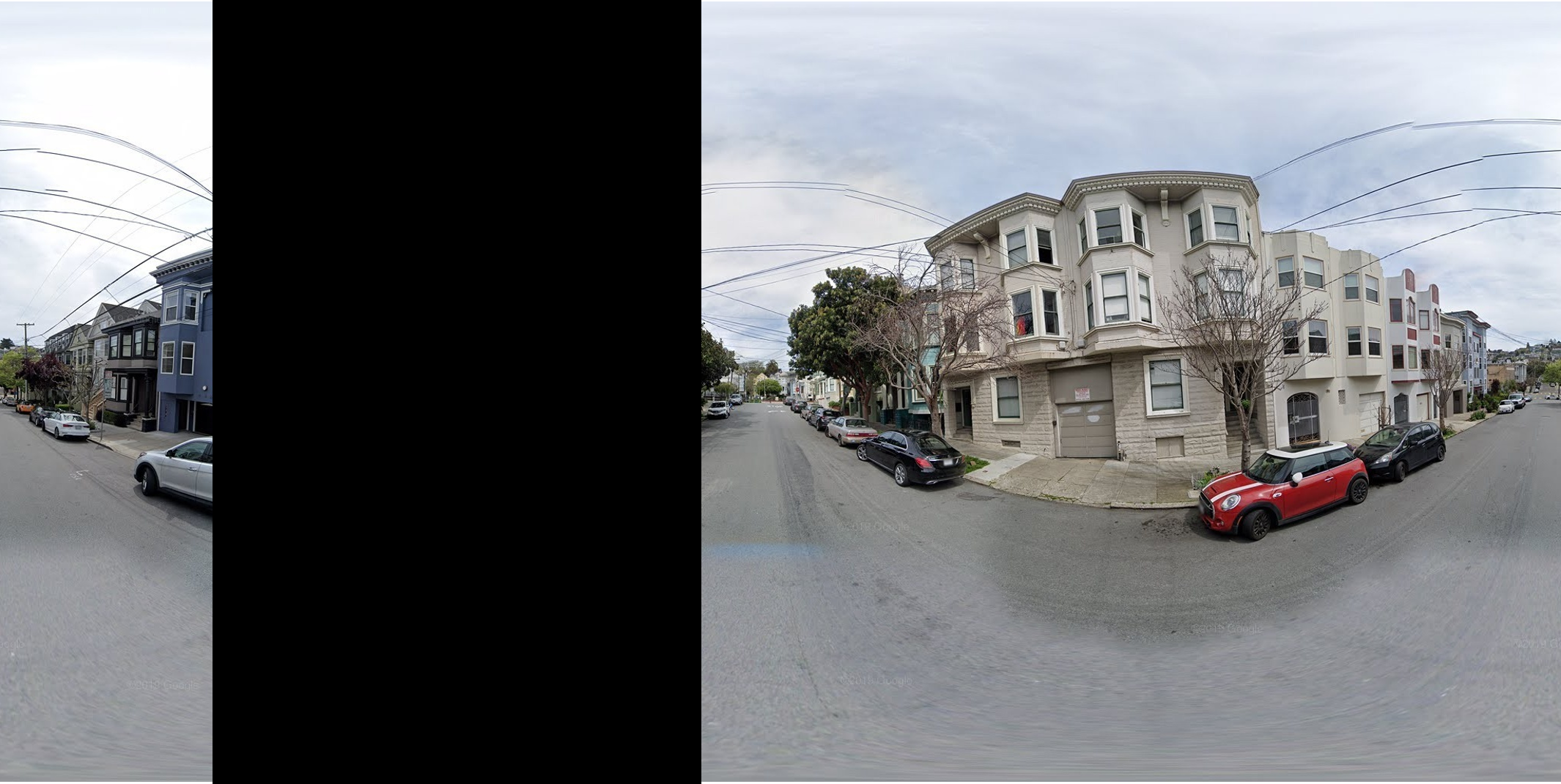}
        \caption{FoV-based Masking}
        \label{subfig:FoVMask}
    \end{subfigure}
    \caption{Different masking strategies.}
    \label{fig: diff mask}
\end{figure}
%  \begin{figure}[h]
%     \centering
%     \includegraphics[width=0.48\textwidth]{geoDistill/figs/mask.pdf}
%     \caption{Different masking strategies.}
%     \label{fig: diff mask}
% \end{figure}

\textbf{Why FoV-based masking?}
% To learn discriminative local features, our GeoDistill proposes FoV-based masking.
% By aligning the predictions from an input with limited information(student) with predictions from an input with full information (teacher), we encourage the student model to focus on learning discriminative local features. 
% Intuitively, random patch-based masking, often used in masked autoencoders~\cite{2021Masked}, and activation-based masking~\cite{2020Self_challenging, devries_improved_2017} which occludes high-activation regions to promote feature decorrelation, can achieve the same goal.
% Since masked autoencoders are typically ViT-based, we benchmarked random masking on both CNN (VGG) and ViT (DINOv2-b14) backbones. The results confirm that our FoV-based masking is superior in both settings, highlighting its backbone-agnostic advantage.
% This is because these two masking strategies risk destroying the image/scene structures and losing important local features, as shown in Fig.~\ref{fig: diff mask}. 
% Consequently, they yield suboptimal performance gains, as demonstrated in Tab.~\ref{tab: diff mask}. 
% In contrast, our approach mimics a query image from a limited-FoV camera. By preserving coherent scene structures, this strategy compels the model to learn more robust and discriminative local features, leading to superior performance.
To learn discriminative local features, our GeoDistill proposes FoV-based masking. 
By aligning predictions from an input with limited information (student) with those from an input with full information (teacher), we encourage the student model to focus on learning discriminative local features. 
Specifically, our approach mimics a query image from a limited-FoV camera. By preserving coherent scene geometry, this strategy compels the model to learn more robust and discriminative local features, leading to superior performance. 
In contrast, alternative strategies such as random patch-based masking, often used in masked autoencoders~\cite{2021Masked}, and activation-based masking~\cite{2020Self_challenging, devries_improved_2017}, risk destroying these crucial scene structures and losing important local features, as illustrated in Fig.\ref{fig: diff mask}. 
This fundamental drawback leads to suboptimal performance gains. To validate this, we performed a comprehensive comparison. 
Given that masked autoencoders are typically ViT-based, we benchmarked random masking against our method on both CNN (VGG) and ViT (DINOv2) backbones. 
Tab.\ref{tab: diff mask} shows that our FoV-based masking is consistently superior in both settings, highlighting its backbone-agnostic advantage.

% making the student model make similar predictions to the teacher model based on an image 
% Our GeoDistill framework employs contiguous region masking to simulate limited FoV, mimicking human visual perception. To assess alternative strategies, we compared it with activation-based~\cite{2020Self_challenging, devries_improved_2017} and random patch masking. Activation-based masking occludes high-activation regions to promote feature decorrelation, while random patch masking follows masked autoencoder (MAE)~\cite{2021Masked}, with two variants: one masking 75\% of the input with 16×16 patches and another matching our random FoV masking ratio (33\%-50\%).

% As shown in Table \ref{tab: mask strategy}, both alternatives significantly underperform our approach, leading to non-convergence of the student network and poorer results. While activation masking aims to enhance generalization and random patch masking leverages image reconstruction techniques, neither proves effective for cross-view localization. This highlights the necessity of preserving spatial context and simulating realistic FoV constraints. Unlike feature dropout, contiguous region masking retains geometric structure, providing a more relevant inductive bias for precise localization and validating our design choice for GeoDistill.

\begin{table}
\centering

\begin{tabular}{llll} 
\hline
             & Data Aug?      & \multicolumn{1}{c}{$\downarrow$Mean(m)} & \multicolumn{1}{c}{$\downarrow$Median(m)}  \\ 
\hline
\multirow{2}{*}{G2SWeakly~\cite{shi2024weakly}} & Yes & 5.64                  & 1.64                    \\
 & No         & \textbf{5.20}                       & \textbf{1.44}                           \\
\hline
\multirow{2}{*}{CCVPE~\cite{xia2023convolutional}} & Yes    & 5.37                  & 2.16                    \\ 
 & No             & \textbf{4.97}                        & \textbf{1.68}                           \\
\hline
\end{tabular}
\caption{Localization performance comparison on two baselines with or without the FoV-based masking as data augmentation on VIGOR~\cite{zhu2021vigor} Cross-Area test set. 
}
\label{tab: data augmentation}
\end{table}

% \textbf{Detrimental Effects of Masking as Data Augmentation}. 
\textbf{FoV-based masking as data augmentation?}
% Due to the extreme perspective disparity between ground and aerial views, compounded by domain gaps in lighting, seasonal variations, the cross-view image matching problem is inherently challenging, even when query images are panoramas with a full $360^\circ$ FoV. 
% When we restrict the FoV of the query images, the difficulty of learning cross-view similarity further increases. 
% Thus, using the FoV-based masking as a data augmentation further increases the learning burden of the model. 
% We demonstrate this in Tab.~\ref{tab: data augmentation}. No matter for supervised learning or weakly supervised learning, using FoV-based masking strategy always impairs the performance. 
% This further demonstrates the necessity of our teacher-student self-distillation pipeline. 
% The FoV-based masking creates a discrepancy between the student (masked input) and the teacher (full input), creating a valuable learning signal, allowing the student to distill knowledge from the teacher and learn robust features from partial views.
In our GeoDistill framework, the proposed FoV-based masking serves as a core component for generating a powerful learning signal. By creating a discrepancy between the student (masked input) and the teacher (full input), it establishes a challenging but valuable self-distillation task. This process forces the student to learn robust features from partial views by distilling knowledge from the teacher's complete perspective, even amidst extreme viewpoint. One might consider applying FoV-based masking as a conventional data augmentation strategy. However, this approach proves detrimental. As demonstrated in Tab.\ref{tab: data augmentation}, simply using FoV-based masking as augmentation consistently impairs performance in both fully and weakly supervised settings. This finding is consistent with results reported in~\cite{xia2023convolutional}, which also found that overly challenging augmentations can degrade model performance. This shows the necessity of our teacher-student pipeline, which successfully transforms this difficult masking task into an effective learning mechanism.

% iven that the FoV masking strategy is integral to GeoDistill, we investigated its effect as a general data augmentation technique in standard training. We applied the same masking approach to the original training pipelines of CCVPE~\cite{xia2023convolutional} and G2SWeakly~\cite{shi2024weakly} baselines, using identical masking ratios. Table ~\ref{tab: data augmentation} shows the localization performance of these baselines with and without FoV masking. Contrary to expectations, we observed a performance drop for both baselines when trained with masking. This result aligns with findings in contrastive learning, such as SimCLR~\cite{chen_simple_2020}, where certain augmentations are detrimental or neutral in supervised learning but beneficial in self-supervised settings. We attribute this performance degradation to the nature of masking, which obscures input information. In standard training, where the goal is to directly map complete inputs to outputs, such information removal hinders optimal feature learning. However, as demonstrated in GeoDistill, this performance gap, created by masked inputs, becomes beneficial when coupled with an appropriate learning objective like geometry-guided self-distillation. The discrepancy between the student (masked input) and teacher (full input) creates a valuable learning signal, allowing the student to distill knowledge from the teacher and learn robust features from partial views.

\label{sec:uncertainty preserved strategy}
\begin{table}
\centering
\begin{tblr}{
  cells = {c},
  hline{1-2,5,6} = {-}{},
    rows = {rowsep=0pt},
    columns = {colsep=18pt},
}
Uncertainty & $\downarrow$Mean(m) & $\downarrow$Median(m)        \\
Single-mode & 4.96          & 1.36          \\
W/o sharpen & 5.23          & 1.44          \\
W/ sharpen     & \textbf{4.49} & \textbf{1.22} \\
Baseline      & 5.20 & 1.44  \\
\end{tblr}
% \setlength{\tabcolsep}{18pt}
% \begin{tabular}{c|c|c}
% \toprule
%     Uncertainty & Mean($\downarrow$)          & Median($\downarrow$)        \\\midrule
% Single-mode & 4.96          & 1.36          \\
% W/o sharpen & 5.23          & 1.44          \\
% Sharpen     & \textbf{4.49} & \textbf{1.22}  \\ \bottomrule
% \end{tabular}
\caption{Localization performance comparison with different uncertainty preservation strategies on VIGOR~\cite{zhu2021vigor} Cross-Area test set. Single-mode preserves the highest confidence prediction.}
\label{tab: uncertainty preservation strategy}
\end{table}
\textbf{Effectiveness of uncertainty preservation strategy}. 
% To validate our uncertainty-preserving sharpening strategy, we conduct an ablation study against two alternatives: single-mode distillation~\cite{xia2024adapting} and a baseline without sharpening (w/o sharpen). Single-mode distillation enforces a deterministic target by selecting only the highest-confidence prediction, discarding the valuable uncertainty information of the teacher model's outputs. The w/o sharpen baseline, which removes temperature sharpening from both student and teacher predictions, fails to converge.
% As Table \ref{tab: uncertainty preservation strategy} shows, our approach significantly outperforms both alternatives, with the variant w/o sharpen highlighting the necessity of sharpening for stable training. By sharpening the teacher’s probability distribution, GeoDistill reduces noise while preserving uncertainty in supervision signals, which is crucial for robust localization. 
A key advantage of our teacher-student framework lies in its ability to distill dark knowledge — the valuable spatial uncertainty contained in the teacher's heatmap. 
We validate this via an ablation study against two extremes, as shown in Table~\ref{tab: uncertainty preservation strategy}. 
On one hand, single-mode distillation~\cite{xia2024adapting} discards this dark knowledge by enforcing a deterministic, single-point target, which leads suboptimal. 
On the other hand, a baseline without sharpening (w/o sharpen) fails to converge because the raw dark knowledge is too diffuse and noisy to serve as a stable learning signal. 
Our approach strikes a crucial balance. By sharpening the teacher's probability distribution, it refines the dark knowledge—filtering out noise while preserving the essential uncertainty. 
This transforms the teacher's output into a potent and stable supervision signal, which is essential for robust localization.
% In contrast, single-mode distillation loses valuable uncertainty information of the teacher model's outputs, and w/o sharpen struggles with convergence. 
% The superior performance and stability of sharpening validate its role in enabling precise and reliable student learning.

% \label{app: fov size}
 
% Thus, we 
% We ablated the input FoV size under the challenging VIGOR cross-area setting to assess its impact on student network localization performance. Testing a range of FoVs revealed that performance remained stable across a considerable range, as shown in \figurename~\ref{fig: diff fov}, but degraded with excessively narrow or wide FoVs.  Specifically, both extremes yielded suboptimal localization accuracy, indicating an optimal FoV range for effective feature extraction.  Motivated by these findings and to introduce bio-inspired variability mimicking human vision, we adopted a dynamic, randomized FoV (180°-240°). This choice, approximating the human eye's FoV,  enhances both biological plausibility and input data randomness, potentially improving generalization and robustness.  Our ablation highlights the importance of FoV selection, suggesting moderate to wide or dynamic FoVs as preferable for optimal localization in our context.

\begin{figure}[t!]
    \centering
    \includegraphics[width=0.45\textwidth]{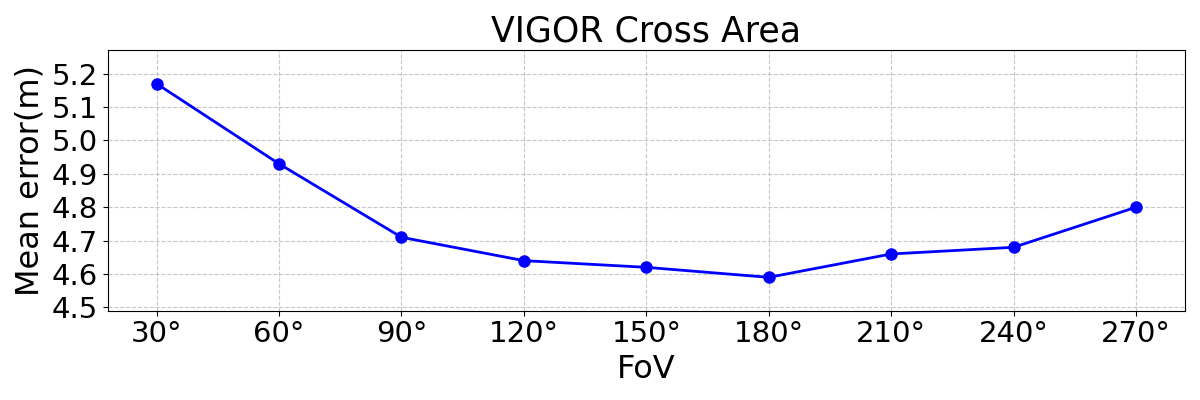}
    \caption{Mean localization error of the student model when trained with different FoVs on VIGOR~\cite{zhu2021vigor} cross-Area test set.}
    \label{fig: diff fov}
\end{figure}
\textbf{FoV size selection}. 
% To determine the appropriate FoV size, we conduct experiments when using different fixed FoVs for training our GeoDistill. The mean localization error is shown in Fig.~\ref{fig: diff fov}. 
% From the figure, the mean localization error becomes large when the retained FoV is too large (over $240^\circ$) or too small (lower than $90^\circ$). 
% This is reasonable as excessively narrow FoV increases the learning burden of the student network, and the performance gap between teacher and student model is small when the student model takes input with a large FoV image, resulting in a weak supervision signal. 
% Motivated by these findings and to introduce bio-inspired variability mimicking human vision, we adopt a dynamic, randomized FoV ($180^\circ\sim240^\circ$). This choice, approximating the human eye's FoV,  enhances both biological plausibility and input data randomness, potentially improving generalization and robustness.  
The choice of FoV for masking is critical to the success of GeoDistill. 
An excessively narrow FoV provides the student with insufficient contextual information, making the learning task intractable. 
Conversely, an overly large FoV makes the student's input too similar to the teacher's full view, diminishing the discrepancy between them and resulting in a weak, ineffective supervision signal.
Fig.~\ref{fig: diff fov} confirms our hypothesis. 
Mean localization error increases significantly when the FoV is either too small (lower than $90^\circ$) or too large (over $240^\circ$).
Motivated by these findings, we adopt a dynamic FoV for training, randomly sampling from the $180^\circ$ to $240^\circ$ range for each instance. 
This design maintains a balance between task difficulty and solvability in distillation, and the incorporation of randomness contributes to improved generalization and robustness.

\section{Conclusion}
\label{sec:conclusion}
We present GeoDistill, a weakly supervised self-distillation framework that improves cross-view localization by learning salient local features. Through teacher-student learning with FoV-based masking, our method enhances localization accuracy and reduces uncertainty, showing significant gains on multiple datasets, particularly those lacking precise annotations. We also introduce a novel orientation network that predicts relative orientation without location supervision, overcoming a key limitation of previous weakly supervised approaches. GeoDistill provides a scalable and effective solution for both weakly and fully supervised methods, demonstrating its high potential for large-scale, real-world localization tasks.

\section{Acknowledgment}
\label{sec:acknowledgement}
The authors are grateful for the valuable comments and suggestions by the reviewers and AC. This work was supported by NSFC (62406194), Shanghai Frontiers Science Center of Human-centered Artificial Intelligence (ShangHAI), MoE Key Laboratory of Intelligent Perception and Human-Machine Collaboration (KLIP-HuMaCo). A part of the experiments of this work were supported by the core facility Platform of Computer Science and Communication, SIST, ShanghaiTech University. 

{
    \small
    \bibliographystyle{ieeenat_fullname}
    \bibliography{main}
}
\clearpage
\appendix

% \clearpage
% \section*{Appendix}

\section{Teacher-student Parameter Update Strategy}
\begin{table}
\centering

\resizebox{\linewidth}{!}{%
\begin{tblr}{
  width = \linewidth,
  colspec = {Q[281]Q[137]Q[175]Q[137]Q[175]},
  cells = {c},
  cell{1}{1} = {r=2}{},
  cell{1}{2} = {c=2}{0.312\linewidth},
  cell{1}{4} = {c=2}{0.312\linewidth},
  hline{1,3,6,7} = {-}{},
  rows = {rowsep=0pt},
}
\begin{tabular}[c]{@{}c@{}}Teacher \\  Param Update\end{tabular}      & Cross-Area    &               & Same-Area     &               \\
             & Mean($\downarrow$)          & Median($\downarrow$)        & Mean($\downarrow$)          & Median($\downarrow$)        \\
Fixed        & 4.65          & 1.28          & 4.46          & 1.43          \\
Prev. Student & 5.02          & 1.37          & 4.49          & 1.44          \\
EMA          & \textbf{4.49} & \textbf{1.22} & \textbf{4.26} & \textbf{1.37} \\
Baseline      & 5.20 & 1.44 & 4.81 & 1.61 \\
\end{tblr}
}
\caption{Localization performance comparison with different update strategies for the teacher network in the VIGOR dataset~\cite{zhu2021vigor}. 
% \textbf{Best in bold}.
Prev Student means using the student from the last epoch as the teacher for the current epoch. Fixed means the teacher network does not update during training. EMA refers to the exponentially moving average teacher adopted in our method.}
\label{tab: different teacher}

\end{table}
We investigate different strategies for updating the teacher model's parameters, including keeping the teacher model's parameters fixed, denoted as ``Fixed'', and using the student model's parameters from the last epoch as the teacher model's parameters for the current epoch, denoted as ``Prev Student''. 
As shown in Tab. ~\ref{tab: different teacher}, all these different teacher parameters update strategy improves the performance over the baseline model, demonstrating the effectiveness of our key idea: using different FoVs to create a discrepancy between teacher and student models, and this discrepancy works effectively as a learning signal to encourage the model focusing on discriminative local features that are useful for cross-view matching. 
Compared to Fixed and our EMA parameters update strategy, Prev. Student suffers from abrupt parameters shifts, which causes significant location prediction inconsistency (before and after teacher parameters update) for some examples, resulting in inconsistent supervision which negatively affects the magnitude of the performance improvement.
In contrast, our EMA update strategy combines the merits of Fixed and Prev. Student. 
It inherits the stability of a fixed teacher model while also adaptively integrating the student’s refined knowledge, resulting in the most considerable performance improvement. 

\section{Different Training Objectives for Self-Distillation}
\label{app: diff distill loss}

To evaluate the impact of different training objectives, we performed an ablation study comparing Cross-Entropy (CE) and Kullback-Leibler Divergence (KLD) as loss functions for our student network. 
Table \ref{tab: diff loss} shows that CE and KLD achieve a similar localization accuracy. 
\begin{table}[h]
\centering

\resizebox{\linewidth}{!}{%
\begin{tblr}{
  width = \linewidth,
  colspec = {Q[302]Q[133]Q[169]Q[133]Q[169]},
  cells = {c},
  cell{1}{1} = {r=2}{},
  cell{1}{2} = {c=2}{0.302\linewidth},
  cell{1}{4} = {c=2}{0.302\linewidth},
  hline{1,3,5} = {-}{},
}
Loss          & Cross-Area &        & Same-Area &        \\
              & Mean($\downarrow$)   & Median($\downarrow$) & Mean($\downarrow$)   & Median($\downarrow$)  \\

KLD & 4.50       & 1.22   & 4.25      & 1.37   \\
CE (ours) & 4.49       & 1.22   & 4.26      & 1.37   
\end{tblr}
}
\caption{Localization performance comparison with different training objective in VIGOR dataset.}
\label{tab: diff loss}
\end{table}

\section{Comparison with Fully Supervised Methods in VIGOR Same Area Test Set}
\label{app: same area comparision}

\begin{table}[ht]
\centering
\caption{Localization performance comparison on VIGOR Same Area test set. \textbf{Best in bold}. \underline{The second-best is underlined}. Here, ``*'' indicates fully supervised methods.}
\label{tab: same area performance}
\resizebox{\linewidth}{!}{%
\begin{tblr}{
  % 列定义
  colspec = {Q[c, wd=4em] l Q[c] Q[c] Q[c] Q[c]}, % 使用 c 模式让 Q 列居中，并给第一列一个参考宽度
  % 全局行/列设置
  rows = {rowsep=0pt},
  row{1,2} = {c, m}, % 表头行居中对齐
  % 单元格合并（这是最佳实践的核心）
  cell{1}{1} = {r=2}{}, % "Noise" 跨2行
  cell{1}{2} = {r=2}{}, % "Method" 跨2行
  cell{1}{3} = {c=2}{}, % "Localization" 跨2列
  cell{1}{5} = {c=2}{}, % "Orientation" 跨2列
  cell{3}{1} = {r=6}{c, m}, % "0°" 跨6行，并垂直居中
  cell{9}{1} = {r=3}{c, m}, % "±45°" 跨3行，并垂直居中
  % 线条定义
  hline{1,12} = {-}{},      % 表格顶部和底部实线
  hline{3} = {-}{},        % 表头下方实线
  hline{8} = {2-6}{-}{},   % 在监督方法和 ours 之间画线
  hline{9} = {-}{},        % 在 0° 和 ±45° 区域之间画线
  hline{11} = {2-6}{-}{},  % 在 ±45° 区域的监督方法和 ours 之间画线
  vline{2} = {3-11}{},     % 在第1和2列之间画线（仅数据区）
  vline{3} = {1-11}{},     % 在第2和3列之间画线
  vline{5} = {1-11}{},     % 在第4和5列之间画线
}
% --- 表格主体（现在非常干净）---
% Row 1-2 (Header)
Noise    & Method      & $\downarrow$Localization &                & $\downarrow$Orientation &                 \\
         &             & Mean                     & Median         & Mean                    & Median          \\
% Row 3-8 (0° block)
0°       & CVR[{\color{blue}39}]*         & 8.82                     & 7.68           & -                       & -               \\
         & SliceMatch[{\color{blue}18}]* & 5.18                     & 2.58           & -                       & -               \\
         & Boosting[{\color{blue}26}]*   & 4.12                     & \underline{1.34} & -                       & -               \\
         & CCVPE[{\color{blue}33}]*      & \underline{3.60}         & 1.36           & 10.59                   & 5.43            \\
         & HC-Net[{\color{blue}31}]*     & \textbf{2.65}            & \textbf{1.17}  & 1.92                    & 1.04            \\
         & GeoDistll(ours)                & 4.26                     & 1.37           & -                       & -               \\
% Row 9-11 (±45° block)
$\pm$45° & CCVPE[{\color{blue}33}]*      & \underline{3.50}         & \underline{1.39} & 10.56                   & 5.96            \\
         & HC-Net[{\color{blue}31}]*     & \textbf{2.70}            & \textbf{1.18}  & \textbf{2.12}           & \textbf{1.04}   \\
         & GeoDistll(ours)                & 4.71                     & 1.48           & \underline{2.90}        & \underline{1.11}  \\
\end{tblr}
}
\end{table}
Here, we supply the comparison of GeoDistill using G2SWeakly as backbone with state-of-the-art fully supervised methods in the Same-Area setting of the VIGOR dataset in Tab.~\ref{tab: same area performance}. 
As anticipated, fully supervised methods generally perform better in the same-area setting than our weakly supervised GeoDistill framework. This is because fully supervised methods are trained with precisely annotated ground truth data within the same geographic area used for testing, allowing them to effectively learn area-specific features and optimize for performance within the training distribution. In contrast, GeoDistill, trained with weakly supervised noisy GPS data and designed for cross-area generalization, is not explicitly optimized for same-area performance.  

For the cross-area evaluation, as highlighted in the main paper, GeoDistill achieves the second-best performance among the compared fully supervised approaches, highlighting its excellent generalization ability compared to fully supervised approaches.

\section{Evaluating GeoDistill with Unlabeled Target Domain Data}
For completeness, we evaluated GeoDistill under the unlabeled target domain data assumption of [{\color{blue}34}]. Intriguingly, retraining CCVPE [{\color{blue}33}] with GeoDistill yielded similar performance using either source (4.05m mean error) or target domain data (3.95m) without GT, aligning with [{\color{blue}34}]'s weakly supervised distillation (3.85m) for domain adapting. However, target domain data availability is often impractical, limiting real-world applicability. Moreover, while [{\color{blue}34}] uses reliable teacher predictions as pseudo-labels for retraining, generalization to truly unseen regions remains a concern.  Like fully supervised methods, target-domain fine-tuning approaches risk performance degradation when encountering new out-of-distribution data. Conversely, GeoDistill's key advantage is achieving significant generalization gains by retraining solely on source domain data. This enables robust generalization to arbitrary unseen cities, offering a more scalable and practical solution. GeoDistill's ability to match target-domain adaptation performance without requiring target data, while ensuring superior generalization, underscores its practical utility and generalization prowess.

\end{document}